\ifdefined\pdfsuppressptexinfo\pdfsuppressptexinfo=-1\relax\fi
\documentclass[sigconf,pbalance]{acmart}

\copyrightyear{2027}
\acmYear{2027}
\acmConference[WSDM '27]{The Twentieth ACM International Conference on Web Search and Data Mining}{February 15--19, 2027}{Hong Kong, China}
\acmBooktitle{Proceedings of the Twentieth ACM International Conference on Web Search and Data Mining (WSDM '27), February 15--19, 2027, Hong Kong, China}
\acmISBN{979-8-4007-3117-4/2027/02}
\hypersetup{keeppdfinfo}

\makeatletter
\@ifpackagelater{hyperxmp}{2024/03/18}{}{\let\wsdm@fixxmp\@empty}
\ifdefined\wsdm@fixxmp
  \patchcmd{\hyxmp@add@simple@lang@ii}{\hyxmp@xmlify{#2}}{\def\hyxmp@xmlified{#2}}{}{}
\fi
\ifdefined\Gread@transgrouptrue
  \AddToHook{cmd/@copyrightpermission/before}{\begingroup\let\Gread@transgrouptrue\Gread@transgroupfalse}
  \AddToHook{cmd/@copyrightpermission/after}{\endgroup}
\fi
\makeatother

\setcopyright{cc}
\setcctype{by}
\acmDOI{10.1145/3862322.3863596}
\author{Pin Ni}
\orcid{0000-0003-4516-1249}
\affiliation{\institution{University of Luxembourg}\city{Luxembourg}\country{Luxembourg}}
\additionalaffiliation{\institution{University College London}\city{London}\country{United Kingdom}}
\email{ni.pin@uni.lu}

\author{Angela Altmeier}
\orcid{0009-0007-5690-7372}
\affiliation{\institution{University of Luxembourg}\city{Luxembourg}\country{Luxembourg}}
\email{angela.altmeier@uni.lu}

\author{Christian Fisch}
\orcid{0000-0003-3609-7193}
\affiliation{\institution{University of Luxembourg}\city{Luxembourg}\country{Luxembourg}}
\email{christian.fisch@uni.lu}

\renewcommand{\shortauthors}{Pin Ni, Angela Altmeier, \& Christian Fisch}

\begin{document}

\title{Certify Before Recalibrating: Exact Support Tests for Label-Shift~EM in Rare Web Streams}

\begin{abstract}
A model can rank cases for review and still misjudge the workload. Do today's scores support enough events to fill a review queue? For binary label-shift expectation maximization (EM) with a nonflat likelihood, one pass through the scores shows whether the fitted event rate exceeds a chosen minimum, or floor. The test evaluates the classical likelihood derivative at that interior floor with scores and source prevalence fixed. From any interior start, EM converges to zero exactly when mean score odds do not exceed source odds. For multiple classes, we construct an improving step when the likelihood favors an excluded class. We tested two predictors, a per-record model and an aggregate rate table, over 30 periods in three streams: the Legal Entity Identifier registry, SEC Form D filings, and Kickstarter. All 17 of the 60 estimates at least 99 percent below the observed rate had an EM optimum at zero. One per-record registry estimate passed the test at a floor of half the anchor, the mean of the last three observed rates, yet was 688 percent too high. In post hoc per-record replays, always anchoring gave lower mean absolute relative error in each stream than switching to EM when the test passed. Both predictors shared the anchor in one 15-day registry holdout. The per-record model's average precision (PR-AUC) was 0.003865, against 0.000785 for the table (4.93-fold). Mean log loss favored the model. The Brier point estimate favored the table. An analyst can test whether EM supports a floor before outcomes arrive and check the estimated count once they do. \space
\end{abstract}



\maketitle
\hypersetup{pdfsubject={Information systems -> Data mining; Computing methodologies -> Supervised learning; Information systems -> Web mining; General and reference -> Empirical studies; }}

\section{Introduction}\label{sec:intro}
\looseness=-1 Web risk models must rank cases and estimate the workload they create. Delayed feedback makes both tasks hard in advertising \cite{chapelle2014delayed,chen2022delayed} and other web streams. Quantification estimates prevalence before labels arrive \cite{moreo2021quapy,schumacher2025comparative}. Under shift, a useful ranking can coexist with a severely wrong event count. We study this separation in three web streams. We track newly published changes such as dissolutions and identifier successions in the Legal Entity Identifier (LEI) registry of companies and other legal entities, repeat SEC Form D filings, and cancellations on the crowdfunding platform Kickstarter.

\begin{figure*}[t]
\centering
\includegraphics[width=0.86\textwidth]{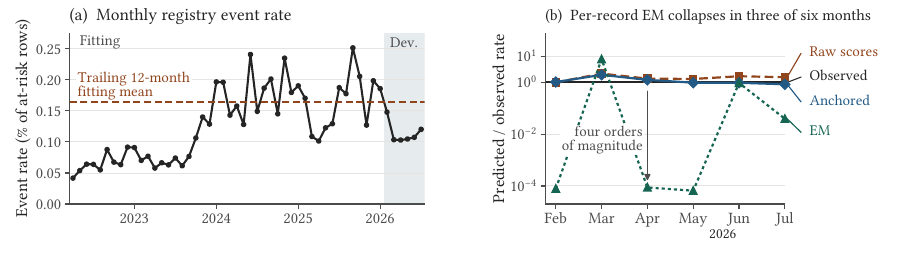}
\caption{Monthly event rates in the LEI registry \cite{gleifgoldencopy}. Per-record EM underestimates the event rate by four orders of magnitude in three months. In March its likelihood has an interior optimum, yet EM predicts nearly eight times the observed rate.}
\Description{Monthly event rates cover fifty-two registry months. They rise from below 0.05 percent to 0.25 percent and then fall. The shaded band marks development and the dashed line gives the mean of the last twelve fitting months. The second panel shows six development months on a log scale. Per-record EM predicts about one ten-thousandth of the observed rate in February, April, and May and about eight times the observed rate in March. Raw scores average above the observed rate in every month, while the anchored predictions stay within a factor of two.}
\label{fig:regime}
\end{figure*}

Label-shift expectation maximization (EM) estimates the current event rate by adjusting source-model probabilities for unlabeled target records \cite{saerens2002adjusting,garg2020unified}. In three of six registry development months, our per-record model still ranked records usefully, but EM stopped four orders of magnitude below the observed rate (Figure~\ref{fig:regime}). More iterations would have driven each estimate closer to zero.

We ask what the deployed scores can establish about the EM update before target labels arrive. An analyst planning for more than $B$ events among $n$ records can set a floor $t=B/n$ before seeing the scores. The slope of the classical concave log-likelihood at $t$ places its optimum above, at, or below that interior floor. At zero, the slope compares the mean score odds with the source odds. For a nonflat likelihood, EM converges to zero from every interior start exactly when their ratio is at most one. The title's ``certify'' refers to this one-sided test of likelihood support.

A registry might distinguish mergers, dissolutions, and bankruptcies. If one model treats them as mutually exclusive classes, can the fit assign one of them zero prevalence? At an optimum on a face of the simplex, the Karush--Kuhn--Tucker (KKT) conditions tell us whether adding an excluded class improves the fit. We construct an improving step when a condition fails and bound the distance to the optimum where the log-likelihood is strongly concave on that face. Our streams have binary outcomes.

We compare the fitted rates of two predictors, one per-record and one aggregate, with observed rates over 30 periods in the three streams. All 17 estimates at least 99 percent below the observed rate had an EM optimum at zero. This match checks our EM code against the likelihood. The two predictors share each period's outcomes, so the 60 estimates are dependent. Probabilistic classify-and-count (PCC), the mean predicted probability, gave lower mean prevalence error than EM in each stream (Section~\ref{sec:replays}).

Even a fitted rate above the floor can be far off. One was 688 percent too high. Earlier labels provide a separate source of information about the rate. We built a registry panel of 130.7 million entity-months. Before holdout scoring, we fixed our per-record hazard model and an aggregate rate table. Both used one anchor, the mean observed rate of the three preceding labeled months. In the 15-day August holdout, the model improved PR-AUC 4.9-fold and lowered log loss. In post hoc checks, both gains held under cluster resampling, in an August 5--15 window, and with isotonic recalibration of the aggregate. Files from the first four days had been parsed for counts, including event counts, before the holdout design was final (Section~\ref{sec:holdout}). On development data, logistic regression gave similar PR-AUC, and refitting the per-record model monthly improved its PR-AUC.

\paragraph{Contributions.}
We test workload floors with the classical likelihood derivative, derive where EM stops near zero at a finite tolerance, and give an improving step for an excluded class. Replays on three streams show how far even an estimate that passes the test can miss the event count.\footnote{Code, reproduction guide, input manifests, and both supplements are at \url{https://github.com/MaxwellNi/fbcert}. The package checks registered numerical claims against its records and regenerates the figures. A full rebuild from the raw archives changed no reported conclusion; the logistic comparator's metrics depend on the processor, within the guide's allowance.}

\section{Finite-Batch Support for Recalibration}\label{sec:regime}
\subsection{Problem setting}
Each entity contributes a row in every period when it is at risk, labeled by whether the event occurs in that period \cite{allison1982discrete}. The ordering and spread of the predicted probabilities describe their \emph{shape}. Their mean gives the predicted event rate, or \emph{level}. Their sum gives the expected count. Observed rates were usually below one percent and reached 1.40 percent on Kickstarter. For period $m$, the replays use current scores and the labels of events dated through $m-1$. Section~\ref{sec:data} describes publication timing.

LeQua 2022 binary test batches contain 250 documents each \cite{esuli2022lequa}. Their smallest possible nonzero prevalence, 0.4 percent, is about eight times our registry holdout rate. We first ask what rate the current scores can support under the fitted likelihood.

\subsection{Exact finite-batch support}\label{sec:boundary}
In the EM update of \citet{saerens2002adjusting}, only the assumed prevalence changes between iterations. The binary condition for an interior mixture optimum \cite{titterington1985statistical,tasche2013art,tasche2014exact,tasche2017fisher}, concavity, convergence, and local rates \cite{peters1976numerical,dempster1977em,wu1983convergence,meng1994rates,alexandari2020maximum} are classical. Population consistency and variance require assumptions on how the scores arise \cite{garg2020unified,tasche2021minimising}. For a nonflat binary likelihood (some score differs from the source prevalence), the derivative at an interior floor $t$ shows whether the fitted rate lies above, at, or below $t$.

\begin{proposition}[Binary support test from likelihood concavity]
\label{prop:emboundary}
For a batch of $n$ records, fix the source prevalence $\pi_s\in(0,1)$ and scores $s_i\in(0,1)$, held fixed across iterations, with $s_i\ne\pi_s$ for at least one $i$. Set $X_i=\frac{1-\pi_s}{\pi_s}\frac{s_i}{1-s_i}$, the relative odds, and $\mu=n^{-1}\sum_iX_i$, their mean. The EM map $F$ and the log-likelihood $L$ (up to an additive constant) are
\begin{equation}\label{eq:em-map-and-likelihood}
F(q)=\frac1n\sum_i\frac{qX_i}{1+q(X_i-1)},\quad
L(q)=\sum_i\log(1+q(X_i-1)).
\end{equation}
$F$ is analytic near $[0,1]$, strictly increasing, and fixes both endpoints. The normalized likelihood derivative
\begin{equation}\label{eq:support-derivative}
C(t)=\frac{L'(t)}n=\frac1n\sum_i\frac{X_i-1}{1+t(X_i-1)}
\end{equation}
is strictly decreasing. For any floor $t\in(0,1)$, the unique maximizer $q^\star$ of $L$ on $[0,1]$ lies above, at, or below $t$ exactly when $C(t)$ is positive, zero, or negative. Also $C(0)=\mu-1$, $F'(0)=\mu$, and
$F(q)-q=q(1-q)C(q)$. From any $q_0\in(0,1)$, the iterates $q_{k+1}=F(q_k)$ converge monotonically to $q^\star$:
(i) if $\mu\le1$, then $q^\star=0$ and $q_{k+1}/q_k\to\mu$;
(ii) if $C(1)\ge0$, then $q^\star=1$;
(iii) if $\mu>1$ and $C(1)<0$, then $q^\star$ is the unique interior zero of $C$, a fixed point of $F$.
\end{proposition}

\begin{proof}
Write $d_i(q)=1+q(X_i-1)\ge\min(1,X_i)>0$. Differentiating gives $F'(q)=n^{-1}\sum_iX_i/d_i(q)^2>0$. Also $L''(q)=-\sum_i(X_i-1)^2/d_i(q)^2<0$. Strict concavity gives the floor test and the three cases. The identity follows from $X_i-d_i(q)=(1-q)(X_i-1)$. Hence the iterate moves toward the maximizer. Monotonicity of $F$ prevents it from crossing an interior fixed point. Continuity identifies its limit, and differentiability gives the ratio at zero.
\end{proof}

For several outcomes, a boundary solution can assign zero weight to a whole group of classes. The next result tests whether bringing one of them back improves the fit.

\begin{proposition}[Multiclass progress, escape, and face distance]
\label{prop:multiclass-progress}
For $K$ classes assume $\pi_{s,k}>0$ and finite nonnegative scores. Put $a_{ik}=s_{ik}/\pi_{s,k}$, $L(q)=\sum_i\log(q^\top a_i)$, and $g_k(q)=\partial_kL(q)$, with $q^\top a_i>0$ wherever derivatives are evaluated. The simplex EM map is $F_k(q)=q_kg_k(q)/n$, with $q^\top g(q)=n$, and
\begin{equation}\label{eq:multiclass-ascent}
L(F(q))-L(q)\ge nD_{\rm KL}(F(q)\Vert q)\ge0.
\end{equation}
Concavity bounds the gap to a global maximizer by
\begin{equation}\label{eq:multiclass-gap}
L(q^\star)-L(q)\le\max_kg_k(q)-n.
\end{equation}
Let $q^A$ maximize $L$ in the relative interior of face $A$, the points $q$ of the simplex with $q_k=0$ for every class $k$ outside $A$. The gradients of the classes in $A$ equal $n$, and $q^A$ is globally optimal exactly when the gradients of all excluded classes are at most $n$. If $\delta_j=g_j(q^A)-n>0$, set
$z_i=a_{ij}/(q^{A\top}a_i)-1$, $V_j=\sum_i z_i^2$, and $\epsilon_j=\min\{1/2,\delta_j/(4V_j)\}$. With $e_j$ the $j$th unit vector, moving to $(1-\epsilon_j)q^A+\epsilon_je_j$ gains at least $\epsilon_j\delta_j-2V_j\epsilon_j^2>0$.

On a face of dimension at least one, let $U$ have orthonormal columns spanning its tangent space, $U^\top U=I$, and put
\begin{equation}\label{eq:face-curvature}
\lambda_0=\lambda_{\min}\!\left(U^\top\!\left[\frac1n\sum_i\frac{a_i a_i^\top}{(\max_{k\in A}a_{ik})^2}\right]U\right).
\end{equation}
Here $\lambda_{\min}$ denotes the smallest eigenvalue. If $\lambda_0>0$, then $\|q^A-q\|_2^2\le2\{\max_{k\in A}g_k(q)/n-1\}/\lambda_0$ for every $q$ on the face where $L$ is finite. If $\lambda_0=0$, $L$ is flat along a tangent direction and $q^A$ is not unique.
\end{proposition}

\begin{proof}[Proof sketch]
Jensen applied to responsibilities yields the ascent inequality \cite{dempster1977em}. The simplex Frank--Wolfe gap \cite{jaggi2013frankwolfe} bounds suboptimality by $g(q)^\top(q^\star-q)\le\max_kg_k(q)-n$ and gives the face KKT test. For the escape, $z_i\ge-1$, so $1+\epsilon z_i\ge1/2$ on $[0,1/2]$. Hence $(\sum_i\log(1+\epsilon z_i))''\ge-4V_j$ without an upper bound on positive $z_i$. Taylor's inequality gives the gain. Since $q^\top a_i\le\max_{k\in A}a_{ik}$ uniformly on the face, the negative Hessian is at least $n\lambda_0$ in orthonormal tangent coordinates. Strong concavity and the gap bound give the distance bound. The mathematical supplement has full derivations.
\end{proof}

If the constrained maximizer has smaller support, take $A$ to be its active face. In the binary case, an exact score of one excludes a zero optimum, and an exact zero excludes a unit optimum. Proposition~\ref{prop:emboundary} treats interior scores. On the classification benchmarks Iris, Wine, and Digits, 15 of 16 class-removal shifts admitted an optimum with zero weight for the removed class. The exception was digit 7. Restoring it increased the likelihood despite its absence from the labels. Conversely, the likelihood can assign zero weight to a class present in the labels when its scores provide too little separation. At equality $g_j=n$, other optima can assign positive weight to $j$ (for example, two classes with identical score columns). EM retains a zero weight once initialized or rounded to zero, even if the gradient identifies an improving step. The benchmark checks and 1,750 randomized trials tested the multiclass implementation.

\begin{corollary}[Finite-tolerance stopping]
\label{cor:stopping}
In Proposition~\ref{prop:emboundary} let $\mu<1$, and stop at the first movement $|q_{k+1}-q_k|<\eta$, returning $q_{k+1}$. Since $0<q_k-q_{k+1}<q_k$, the loop stops no later than the first $q_k<\eta$. As $\eta\downarrow0$, $F(q)=\mu q+O(q^2)$ bounds the returned value between $(\mu^2+o(1))\eta/(1-\mu)$ and $(\mu+o(1))\eta/(1-\mu)$. One more update multiplies it by $\mu+o(1)$. On a linear trajectory $q_k=q_0\mu^k$, the strict rule takes $\max(1,2+\lfloor r\rfloor)$ updates, where $r=\log(\eta/[q_0(1-\mu)])/\log\mu$. For EM this count is approximate because of the nonlinear transient. When $\mu=1$, the leading movement is quadratic and the stopping scale is proportional to $\sqrt\eta$.
\end{corollary}

\paragraph{Reading the test.}
Since $F(t)-t=t(1-t)C(t)$, one EM step from the floor has the sign of the test. At $t=\pi_s$ that step returns the mean score \cite[cf.\ Cor.~7]{tasche2014exact}. Evaluating $C(t)$ gives the comparison directly, without running EM to convergence or locating the maximum by bisection. At zero it suffices to average the score odds and compare with the source odds. Individual records' contributions to this mean reveal which scores drive the result. Under the support and integrability conditions in the mathematical supplement, exact source posteriors and label shift imply that the population mean of the relative odds is at least one. A value below one in a large batch can therefore reflect sampling variation or a violated assumption.

For the observed scores and the chosen $\pi_s$, the sign of $\mu-1$ is exact. Holding the scores fixed gives
$\mu(\pi_s')=\mu(\pi_s)\operatorname{odds}(\pi_s)/\operatorname{odds}(\pi_s')$. Multiplying the source odds by $\mu$ brings $\mu(\pi_s')$ to one, where zero still maximizes the likelihood. Source odds above $\mu\,\operatorname{odds}(\pi_s)$ make zero attractive. Lower source odds make it repelling. Extending this finite-batch sign to future batches requires accounting for score dependence and uncertainty in the source prior $\pi_s$. Interpreting the fitted rate as a population prevalence further requires label-shift and calibration assumptions or more labels. \space

\section{Related Work}\label{sec:related}
\paragraph{Label-free prevalence estimation.}
\citet{saerens2002adjusting} popularized fixed-score EM for classifier outputs, and \citet{esuli2021critical} reassessed it. \citet{peters1976numerical} had earlier derived the iteration from the likelihood gradient. For fixed mixture components, \citet{tasche2014exact} gives the gradient in density-ratio form, writes the ratio as score odds over source odds, and credits the condition for a binary interior optimum to \citet{titterington1985statistical}. The mean ratio and mean inverse ratio must both exceed one. For a nonflat binary likelihood and an interior start, failure of this condition leads EM to a boundary, sometimes slowly. \citet{tasche2017fisher} gives the sample likelihood derivative at a fixed prior and recommends solving its binary root directly. We use the derivative to test a floor and quantify stopping error. \citet{alexandari2020maximum} establish concavity and recommend the mean calibrated prediction on held-out source data as the source prior. Section~\ref{sec:scope} tests this rule. Alternatives to EM include black-box shift estimation (BBSE) \cite{lipton2018detecting} and regularized learning under label shift (RLLS) \cite{azizzadenesheli2019regularized}. \citet{garg2020unified} study population identification, and \citet{tasche2021minimising} studies variance.

Population mixture-proportion estimation studies irreducibility, convergence rates, distribution embeddings, and positive-unlabeled recovery \cite{blanchard2010novelty,scott2015rate,ramaswamy2016mixture,jain2016classprior,bekker2020pu,zhu2023mixture}. \citet{hosseini2026regions} tests population prevalence thresholds using held-out class-conditional reference samples. Our floor test concerns the maximizer of the observed scores' likelihood.

Quantification ranges from adjusted counting and calibrated aggregation \cite{forman2005counting,forman2006quantifying,forman2008quantifying,bella2010quantification,card2018importance} to distribution matching and shared benchmarks \cite{gonzalez2013hellinger,maletzke2019dys,esuli2022lequa,esuli2024lequa,moreo2021quapy}. Recent estimators use semiparametric moments, calibrated networks, distributed likelihood, graph priors, and compositional geometry \cite{tian2023elsa,wen2024cpmcn,wu2025vrls,kimura2025gsb3se,moreo2026geometry}. When class-conditional distributions also change, label-shift prevalence estimates can degrade \cite{gonzalez2024binary,garg2023rlsbench}. Extreme imbalance can favor prior shrinkage \cite{ye2024label}.

\paragraph{Online label shift and delayed outcomes.}
Online methods address unlabeled test loss, moving priors, dynamic regret, and finite-sample calibration \cite{wu2021online,bai2022adapting,baby2023online,jeong2026calibration}. Related work gives conformal uncertainty under label shift and evaluation under delayed verification \cite{podkopaev2021distribution,grzenda2020delayed}. \citet{wu2021online} update from individual unlabeled predictions, whereas our test uses a single batch of scores. \citet{baby2023online} track changing priors from unlabeled data or observed labels. Our anchor averages the observed rates from the three preceding labeled periods. \citet{menon2021longtail} use training-class frequencies to adjust logits in long-tail learning.

\paragraph{Maintaining calibration under drift.}
Venn--Abers uses isotonic regression to calibrate ranking scores \cite{vovk2014venn}. \citet{murphy1973new} decomposes the Brier score into reliability, resolution, and uncertainty. CORP \cite{dimitriadis2021stable} yields consistent, optimally binned, reproducible reliability diagrams with the pool-adjacent-violators algorithm and decomposes proper scores into miscalibration, discrimination, and uncertainty (Table~\ref{tab:holdout-calibration}). Clinical work advocates continual updating \cite{jenkins2021continual} and compares recalibration strategies as outcomes arrive \cite{tanner2023dynamic,schnellinger2021comparison}. Banking similarly rescales default scores to a new base rate \cite{tasche2013art}. Other online methods derive risk certificates under temporal shift \cite{huang2026certificate} or, in multicalibration, target groupwise reliability under adversarial feedback \cite{ghuge2025multicalibration}. We locate the likelihood maximum relative to a floor before labels arrive.

\paragraph{Rare-event registry forecasting.}
\citet{midthune2005modeling} model reporting delays and corrections within strata in cancer registries. Our aggregate rate table stratifies by jurisdiction and legal form. Corporate-default models use industry and market covariates for listed firms \cite{shumway2001forecasting,chava2004bankruptcy,campbell2008search}. We cover registered entities broadly and evaluate level drift after fitting. \space

\section{Three Rare-Event Streams}\label{sec:data}
Of the three streams, only the registry supplies a holdout with predictors and criteria fixed before scoring (Section~\ref{sec:holdout}).

\subsection{The registry panel}
The Global Legal Entity Identifier Foundation (GLEIF) publishes the LEI registry three times daily under CC0. Full snapshots and eight-hour updates cover entity (lei2) and relationship (rr) records \cite{gleifgoldencopy}. We recorded each URL, timestamp, size, and SHA-256, checked coverage against the publisher index, and rehashed a sample (empirical supplement). We replayed updates from the 2022-03-30 baseline to the 2026-08-01 cutoff (Table~\ref{tab:panel}).

At the month's start, each primary row requires ACTIVE status, a registration status outside MERGED, RETIRED, ANNULLED, DUPLICATE, and TRANSFERRED, and no earlier study event. A row is positive if any of four first changes occurs: a completed lineage event, a transition to INACTIVE, a registration newly marked MERGED/RETIRED/ANNULLED/DUPLICATE, or a new successor identifier. Coinciding changes count once. The 14 lineage types include dissolution, liquidation, bankruptcy, demerger, spinoff, and branch conversion. We date each event by its first publication in the replay, including backdated changes. ``First'' refers to the study history. Recurrent events enter only secondary analyses.

\begin{table}[htbp]
\caption{Construction of the registry panel. Event rates use at-risk entity-months as the denominator.}
\label{tab:panel}
\footnotesize
\setlength{\tabcolsep}{4pt}
\begin{tabular}{lr}
\toprule
Files (lei2 and rr), acquired / before cutoff & 9,600 / 9,510 \\
Baseline full-file records (2022-03-30) & 2,101,922 \\
Entity delta files after baseline / parse errors & 4,754 / 0 \\
Records tracked & 3,390,929 \\
Panel rows, 2022-04 to 2026-07 & 130,693,379 \\
Monthly event rate range (percent) & 0.042 to 0.251 \\
Fitting rows / events, 2022-04 to 2026-01 & 112,519,203 / 149,921 \\
Development rows / events, 2026-02 to 2026-07 & 18,174,176 / 20,773 \\
Single-use holdout window & 2026-08-01 to 2026-08-15 \\
\bottomrule
\end{tabular}
\end{table}

The initial comparison used February--July 2026 (Section~\ref{sec:ranking-and-workload}). We also used these months for development. August outcomes did not enter fitting or predictor selection. The fixed 22 feature groups (30 columns) cover record state, ages and clocks, event history, announced events (advance notice of a lineage event), registry context, and parent/child counts. All were available before scoring. Section~\ref{sec:ranking-and-workload} checks ranking on rows with no announced event.

\subsection{Two additional streams}
Issuers file Form D with the U.S. Securities and Exchange Commission (SEC) to give notice of certain exempt securities offerings. We study repeat filings by 339,577 issuers in 72 quarterly archives spanning 2008Q1--2025Q4 \cite{secformddata}. An issuer enters the month after its first filing and remains at risk. Any filing makes that issuer-month positive. The twelve scored months of 2025 have event rates of 0.56--0.85 percent. Because the files are released quarterly, the entire preceding month need not have been public at scoring time.

On the crowdfunding platform Kickstarter, we study campaign cancellation rather than funding success \cite{etter2013launch,mollick2014dynamics}, using 201,434 campaigns from the 2026-06-11 Web Robots crawl \cite{webrobotskickstarter}. A campaign is at risk if it has launched by the start of the Monday-aligned week and has not yet ended. Elapsed-week features use campaign dates. The June crawl supplies attributes whose historical values we cannot verify. In 8,522 of 9,172 canceled campaigns, the crawled title carries a post-outcome ``(Canceled)'' suffix. We strip one copy before measuring title length. Removing repeated copies gave identical predictions (empirical supplement). Twelve scored weeks have 1,423--2,318 at-risk campaigns and 9--22 cancellations (0.39--1.40 percent). Under independence, the relative binomial standard error is 21--33 percent. \space

\section{When Better Ranking Misstates Workload}\label{sec:ranking-and-workload}
We first compared the raw per-record model and the aggregate rate table on the same six registry months.

\subsection{Evaluation protocol}
We fixed the criteria before the initial comparison. At least 2 percent of distinct at-risk entities had to experience an event during fitting, with a Wilson lower bound above 1 percent. The model then needed higher PR-AUC and lower 20-bin equal-count expected calibration error (ECE) than the aggregate, each with a favorable 2,000-draw 95 percent bootstrap interval. Throughout, PR-AUC is threshold-grouped, tie-correct average precision.

The per-record model uses LightGBM 4.6.0 \cite{ke2017lightgbm} with 30 fixed columns and validation from November 2025 to January 2026. Selection by log loss over a small grid at learning rate 0.05 (empirical supplement) gave 31 leaves, unit positive weight, and 93 trees. We refit on the full fitting window. The aggregate crosses 21 jurisdictions with 11 legal forms, including catch-all groups. Cells with fewer than 50 fitting events fall back hierarchically. The rate for month $m$ comes from the same cell a year earlier ($m-12$).

\subsection{Prespecified decision}
The registry met the minimum event-rate requirement. During fitting, 149,921 of 3,074,124 distinct entities experienced an event (4.8769 percent, Wilson lower bound 4.8528 percent). Table~\ref{tab:ranking-and-calibration} reports both criteria on the 18,174,176 development rows (20,773 positives). The bootstrap resampled their 3,116,907 entities.

\begin{table}[htbp]
\caption{Initial registry comparison using raw predictions. The joint criterion required higher PR-AUC and lower calibration error. Margins are per-record minus aggregate.}
\label{tab:ranking-and-calibration}
\footnotesize
\setlength{\tabcolsep}{4pt}
\begin{tabular}{lrr}
\toprule
 & PR-AUC (higher) & ECE, 20 bins (lower) \\
\midrule
Model & 0.007400 & 0.000585 \\
Aggregate & 0.001956 & 0.000341 \\
Margin & +0.005444 & +0.000244 \\
95\% CI & [+0.005185, +0.005718] & [+0.000223, +0.000263] \\
Criterion met? & Yes (3.78-fold) & No \\
\bottomrule
\end{tabular}
\end{table}

The model's PR-AUC was 3.78 times the aggregate's, but its ECE was higher, and both intervals excluded zero. It met only the ranking criterion, and the protocol required both.

\subsection{Monthly rates explain the calibration error}
From March through July, the raw model overpredicts the event rate by 36 to 118 percent (Table~\ref{tab:monthly-rates}).

\begin{table}[htbp]
\caption{Monthly registry event rates and raw predictions before recalibration. Rates are percentages of at-risk rows, and bias is relative to the observed rate.}
\label{tab:monthly-rates}
\footnotesize
\setlength{\tabcolsep}{4pt}
\begin{tabular}{lrrrrr}
\toprule
Month & Events & Observed & Model mean & Model bias & Aggr.\ bias \\
\midrule
2026-02 & 4,373 & 0.1479 & 0.1481 & +0.1\% & +12.4\% \\
2026-03 & 3,089 & 0.1035 & 0.2259 & +118.4\% & +2.7\% \\
2026-04 & 3,106 & 0.1029 & 0.1443 & +40.2\% & $-$3.9\% \\
2026-05 & 3,184 & 0.1046 & 0.1418 & +35.6\% & +14.9\% \\
2026-06 & 3,295 & 0.1073 & 0.1856 & +73.0\% & +18.1\% \\
2026-07 & 3,726 & 0.1202 & 0.1906 & +58.6\% & +53.0\% \\
\bottomrule
\end{tabular}
\end{table}

In each of those months, the aggregate's seasonal profile is closer to the observed rate. Every equal-count bin overpredicts, so the model's ECE equals its pooled level error. Its row-weighted monthly level error exceeds the aggregate's by 0.000376, more than the whole ECE margin.

On the 99.72 percent of rows without an announced lineage event, the PR-AUC ratio drops from 3.78 to 2.23 (0.003967 against 0.001782, paired-difference interval [+0.002085, +0.002301]). The model still ranks better and its ECE remains higher. We neither removed features nor refit the model. Two prespecified secondary analyses, a recurrent-network comparator (PR-AUC 0.006311, ECE 0.003240) and a per-event-type breakdown, leave the decision intact.

Before scoring the holdout, we specified the predictors and the PR-AUC and log-loss criteria (Section~\ref{sec:holdout}). The initial comparison remains a development result. \space

\section{A Delayed-Label Anchor}\label{sec:model}
\subsection{Registry predictions before labels arrive}
\begin{figure}[htbp]
\centering
\includegraphics[width=\linewidth]{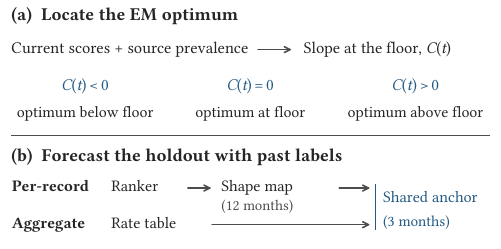}
\caption{Two uses of prevalence information. The slope of a nonflat likelihood at a chosen floor locates the EM optimum relative to that floor. The holdout compares two predictors with the same anchor.}
\Description{The upper panel uses current scores and source prevalence. At an interior floor, a negative, zero, or positive likelihood slope places the optimum below, at, or above the floor. The lower panel shows the separate holdout comparison. A ranker fitted on the fitting window uses a shape map learned on the last twelve labeled months. The aggregate rate table omits this stage. Both predictors use the event rate averaged over the last three labeled months as their common target prior before the period.}
\label{fig:method}
\end{figure}

The likelihood test uses current scores (Figure~\ref{fig:method}). Both holdout predictors take their target prior from past labels. We refit the per-record LightGBM ranker of Section~\ref{sec:ranking-and-workload} on its original fitting window with seed 20260806 (empirical supplement). We fit a monotone shape map, then rescale odds using the target prior.

For each scored month, we fit grid-approximated Venn--Abers isotonic pairs \cite{vovk2014venn,vovk2015large} on the calibration slice, the preceding twelve months of available labels. Assigning a provisional label of zero or one to the scored case gives predictions $p_0$ and $p_1$, combined as $p=p_1/(1-p_0+p_1)$. The nondecreasing map can change rankings only by creating ties, which can change PR-AUC. Drift can violate the exchangeability required for Venn validity. The calibration slice overlaps the ranker's fitting window, completely in the first scored month, so its scores can be in-sample.

The level layer multiplies every record's odds within a scored month by one positive factor, as in credit-risk recalibration under an invariant likelihood ratio \cite{tasche2013art}. With source prior $\pi_s$ (the slice's observed rate for the per-record model, the carried seasonal level for the aggregate) and target prior $\pi_t$, each probability $p$ is mapped to
\begin{equation}\label{eq:level-layer}
p' \;=\; \frac{r_1\,p}{\,r_1\,p + r_0\,(1-p)\,},
\qquad
r_1 = \frac{\pi_t}{\pi_s},
\quad
r_0 = \frac{1-\pi_t}{1-\pi_s},
\end{equation}
which preserves ranking within the month. The mean rescaled probability can differ from the target prior. The level depends on whether $\pi_t$ comes from current scores (Section~\ref{sec:regime}) or past labels.

\subsection{From estimating the level to anchoring it}
On development data we first estimated the rate by EM on each month's unlabeled scores \cite{saerens2002adjusting}, starting at the fitting-window prior. After this EM collapsed (Section~\ref{sec:registry-boundary}), we replaced it with the anchor, the mean observed rate of the three most recent labeled months. Both holdout predictors always used the anchor. We evaluated the gated rule (Section~\ref{sec:scope}) post hoc on development data.

The number of past observations and their correlation determine the anchor's precision. The anchor can also lag when the rate changes (Section~\ref{sec:replays}).

The aggregate uses the same anchor without a Venn--Abers layer. Its scores are constant within each cell, so they cannot order records inside it, although an isotonic map may improve log loss (Sections~\ref{sec:replays} and~\ref{sec:holdout}). Section~\ref{sec:replays} adds a logistic regression model and a monthly refitted ranker on development data only.

\subsection{Holdout design}
The initial comparison used binned ECE. The holdout evaluated individual probabilities with mean log loss, a proper score. Registry updates published from 2026-08-01 through 2026-08-15 supplied the holdout labels. The exposure transform $p_{15}=1-(1-p_{\text{month}})^{15/31}$ assumes a homogeneous within-month hazard and applies to both predictors. The two criteria required higher PR-AUC and lower mean log loss for the model than for the aggregate, each with a favorable 95 percent percentile interval excluding zero (2,000 paired record-bootstrap draws, seed 20260806). Neither set a minimum effect size. July's rate, scaled by 15/31, implied roughly 1,800 positives. \space

\section{EM Collapse Across Streams and Stress Tests}\label{sec:replays}
We replayed the development data post hoc. These analyses leave the prespecified holdout predictors and criteria intact.

\subsection{Boundary behavior in the registry}\label{sec:registry-boundary}
Three of six registry months had per-record EM estimates approaching zero. A fourth underestimated the rate by 96 percent. The other two errors were plus 688 and minus~5 percent (Table~\ref{tab:emcollapse}). Aggregate EM underestimated the rate by 99.94 to 99.97 percent every month.

\begin{table}[htbp]
\caption{Per-record registry EM estimates in six development months. EM mean is the mean after one further update from the stopped prior. Bias is relative to the observed rate. The iteration heuristic uses $\mu=F'(0)<1$ (Corollary~\ref{cor:stopping}). The cap is 1,000.}
\label{tab:emcollapse}
\footnotesize
\setlength{\tabcolsep}{4pt}
\begin{tabular}{lrrrrrr}
\toprule
 & \multicolumn{2}{c}{Rate (\%)} & & & \multicolumn{2}{c}{Iterations} \\
\cmidrule(lr){2-3}\cmidrule(lr){6-7}
Month & Observed & EM mean & Bias & $\mu$ & Obs. & Heur. \\
\midrule
2026-02 & 0.1479 & $1.16\times 10^{-5}$ & $-$99.99\% & 0.92702 & 117 & 123 \\
2026-03 & 0.1035 & $8.15\times 10^{-1}$ & +687.9\% & 1.25794 & 112 & n/a \\
2026-04 & 0.1029 & $8.95\times 10^{-6}$ & $-$99.99\% & 0.90859 & 96 & 100 \\
2026-05 & 0.1046 & $6.82\times 10^{-6}$ & $-$99.99\% & 0.88833 & 80 & 83 \\
2026-06 & 0.1073 & $1.02\times 10^{-1}$ & $-$5.0\% & 1.03595 & 236 & n/a \\
2026-07 & 0.1202 & $4.73\times 10^{-3}$ & $-$96.1\% & 1.00145 & 1,000 & n/a \\
\bottomrule
\end{tabular}
\end{table}

EM uses the per-record Venn--Abers scores or the aggregate hazards. Only the hazards are clipped to $[10^{-12},1-10^{-12}]$ before the support calculation. Log loss clips final predictions separately. The per-record source prior is the calibration slice's observed rate. Section~\ref{sec:scope} tests the mean-prediction rule of \citet{alexandari2020maximum} on that slice and on held-out months.

For the per-record model, only February, April, and May have $\mu$ below one. Their likelihoods peak at zero.

In those three months at tolerance $\eta=10^{-8}$, each EM mean, one update past its returned prior, is within 3.6 percent of $\mu^2\eta/(1-\mu)$. The heuristic closely predicts the observed registry iterations (Table~\ref{tab:emcollapse}) but overpredicts by 14 to 93 percent on six Form~D aggregate pairs (mathematical supplement).

In February the geometric mean factor per EM update was 0.92377, close to $\mu=0.92702$. July's $\mu$ is 1.00145, so zero repels. Its interior maximizer is $3.58\times10^{-5}$ (0.00358 percent), with a limiting bias of minus 97.0 percent versus the capped run's minus 96.1 percent. Record resampling leaves the sign of July's $\mu-1$ unresolved. The analytic 95 percent interval for $\mu$ is [0.9943, 1.0086], and 32.45 percent of 2,000 bootstrap resamples fall below one.

\subsection{Finite-batch results across three streams}
\begin{figure}[htbp]
\centering
\includegraphics[width=0.90\linewidth]{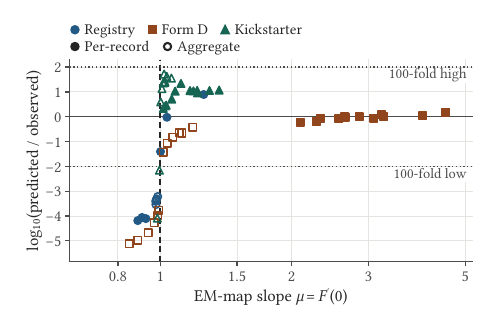}
\caption{Mean predictions after EM stops, relative to observed rates, for 60 predictor-period pairs. All 17 underestimates of at least 99 percent have $\mu<1$. The largest overestimate is 52-fold (Kickstarter aggregate, March 23, 2026).}
\Description{For these 60 predictor-period pairs, shapes distinguish the three streams: circles for the registry, squares for Form D, and triangles for Kickstarter. Filled markers denote per-record predictors and open markers aggregates. The vertical line marks mean relative odds of one. The signed log rate ratio has guides at minus two and plus two, a hundredfold under- or overestimate. All seventeen points at or below minus two lie left of the vertical line. Every point to its right falls between the guides. The largest overestimate is a 52-fold Kickstarter aggregate prediction.}
\label{fig:crossplot}
\end{figure}

The 60 predictor-period pairs share 30 sets of outcomes. Of the 17 estimates at least 99 percent too low, nine were registry cases, six Form D, and two Kickstarter. All three per-record cases were registry months. The other 14 were aggregate. All 17 likelihoods peak at zero (Figure~\ref{fig:crossplot}). Even with an interior optimum, errors reach minus 96.07 percent for the July registry per-record model and minus 96.25 percent for the May 2025 Form~D aggregate.

Per-record EM bias ranged from minus 39 to plus 52 percent on Form D and plus 115 to plus 2,090 percent on Kickstarter, with $\mu$ above one throughout. Aggregate EM bias ranged from minus 100 to plus 5,132 percent. Across the 30 periods, the anchored per-record model's largest absolute biases were 17.9, 72.7, and 96.2 percent on Form~D, Kickstarter, and the registry, respectively.

Table~\ref{tab:quantifiers} uses per-record scores and earlier source periods, at least two on the registry and three each on Form D and Kickstarter. The 22 targets are April--July on the registry, April--December on Form D, and January 26--March 23 on Kickstarter. The latter two streams may use inputs unavailable at scoring time (Section~\ref{sec:data}).

PCC has lower mean error than EM on all three streams. Among methods using labeled-source moments, BBSE leads on Form D and RLLS on the other two. Among methods using earlier target labels, the anchor leads on Form D and Kickstarter and Prior on the registry. Different shifts can change these rankings \cite{gonzalez2024binary,schumacher2025comparative}.

\begin{table}[t]
\caption{Mean absolute relative prevalence error (\%) over 22 equally weighted periods (4 registry, 9 Form D, 9 Kickstarter). PCC averages target scores. EM adds a source prior. BBSE uses hard source counts and RLLS soft score moments. Prior averages all eligible earlier period rates equally. Anchor averages the last three rates.}
\label{tab:quantifiers}
\footnotesize
\setlength{\tabcolsep}{4.5pt}
\begin{tabular}{lrrrrrr}
\toprule
Stream & PCC & EM & BBSE & RLLS & Prior & Anchor \\
\midrule
Form D & 4.10 & 16.26 & 20.60 & 28.56 & 11.77 & 8.62 \\
Registry & 35.64 & 75.26 & 10,657 & 11.85 & 11.93 & 17.64 \\
Kickstarter & 42.31 & 785.96 & 238.74 & 42.45 & 46.26 & 24.82 \\
\bottomrule
\end{tabular}
\end{table}

The empirical supplement compares adjusted classify-and-count (ACC), its probabilistic version (PACC), and maximum-likelihood label shift (MLLS). All 208 BBSE and RLLS prevalence estimates from published implementations agreed with independent calculations at the stated tolerances.

Registry hard BBSE has 10,657 percent mean error (monthly range 5,753 to 18,437 percent). Its source matrices cross predicted class $\mathbf1[p>0.5]$ with true class and divide counts by source sample size. All four are full rank but ill-conditioned, with condition numbers of $1.44\times10^7$ to $1.72\times10^7$. Hard RLLS gives 11.86 percent error and soft RLLS 11.85. Their weights stay within $2.5\times10^{-4}$ of one, so both nearly return the pooled source rate.

\subsection{Choosing a floor and quantifying uncertainty}\label{sec:scope}
A $\mu$ just above one says little about the error. Three pairs with $\mu$ near 1.04, one from each stream, had errors of minus 92, minus 5, and plus 4,138 percent. The registry's July per-record optimum is 0.00358 percent, so $C(t)<0$ for every higher floor despite $C(0)>0$. The test is tail-sensitive. Dropping the highest-scoring 0.01 percent of registry rows moves June's $\mu$ from 1.036 to 0.993 and July's from 1.001 to 0.962. In the per-record Kickstarter week of February~16, winsorizing the top 1 percent of scores moves $\mu$ from 1.016 to 0.996. At the floor $\pi_s$ (the mean-score comparison), the fitted optimum falls below the floor in all 17 collapses and in 17 more pairs with optima in $(0,\pi_s)$.

In a post hoc replay, we compared floor decisions and uncertainty estimates on all 60 batches. Pointwise 95 percent Poisson-multiplier bootstrap intervals for $\mu$ (400 draws) contain one for 10 batches: one registry, three Form D, and six Kickstarter (11 with normal intervals). We resampled 21 jurisdictions or 40 managing Local Operating Units (LOUs), keeping the prior and shape map fixed. All 12 registry intervals contain one for each partition (2,000 draws each). Each batch's sign of $\mu-1$ is exact, but cluster resampling leaves the registry population signs unresolved. A two-sample Kolmogorov--Smirnov test against the calibration slice flags all 30 per-record batches, collapsed or not (empirical supplement). At half the anchor, the original stopped EM agrees with every exact floor decision. A coarse run (tolerance $10^{-4}$, at most 100 steps) disagrees on 12 of 60, eight after a single step. The observed rate exceeded the floor in every batch. Of the fitted rates, 24 fell below the floor and 36 exceeded it. Of those 36, 24 missed the observed rate by more than 50 percent.

\begin{table}[htbp]
\caption{Post hoc mean absolute relative error (\%) of the final predicted mean over 30 per-record periods. Table~\ref{tab:quantifiers} compares prevalence estimates on 22 periods. Gated uses the EM optimum if $C(\bar{\pi}/2)>0$ and otherwise the anchor $\bar{\pi}$, the mean of the three preceding observed rates. EM uses its stopped prior. Each chosen prior rescales row odds before averaging. No update averages calibrated scores.}
\label{tab:policies}
\footnotesize
\setlength{\tabcolsep}{4.5pt}
\begin{tabular}{lrrrrr}
\toprule
Stream & Periods & EM & Anchor & No update & Gated \\
\midrule
Registry & 6 & 181.49 & 24.24 & 37.29 & 123.12 \\
Form D & 12 & 17.74 & 7.45 & 4.86 & 17.75 \\
Kickstarter & 12 & 832.71 & 29.77 & 35.60 & 832.72 \\
\bottomrule
\end{tabular}
\end{table}

When the fitted rate exceeded half the anchor, the gated rule used the EM optimum. Otherwise it used the anchor. It avoided all three per-record collapses but kept March's estimate, 688 percent too high. Always anchoring or keeping the calibrated scores gave lower mean error than the gated rule in every stream (Table~\ref{tab:policies}). Always anchoring also gave lower median error and log loss. Registry log loss was 0.009081 with the gated rule and 0.008227 with anchoring. The rule selected EM in 26 of 30 periods, and the anchor was closer in 23 of those 26. Floors at one quarter or one tenth of the anchor gave the same choices. We examined all these floors after seeing the scores and outcomes.

\paragraph{Source-prior sensitivity.}
The recommended mean-prediction prior does not remove the zero optima. On the replay's scores, we set $\pi_s$ to the calibration slice's mean calibrated score, within 1.2 percent of its observed rate on the registry. The per-record registry model kept its February, April, and May zero optima and gained one in July. Kickstarter gained five zero-optimum weeks and Form~D none. A stricter April--July registry check withholds source month $m-1$'s labels from both ranker and shape-map fitting. On identical target scores, we compare that month's label rate with its mean calibrated probability \cite{alexandari2020maximum}. Unlabeled source and target scores enter the Venn--Abers quantile grid. The mean-prediction prior gives zero optima in April, May, and July and a plus 1,115 percent error in June. The label-rate prior gives no zero optimum but errors of plus 1,024 to plus 18,177 percent.

\subsection{Controlled violations of label shift}\label{sec:synthetic-shift}
We compared likelihood support with truth in a post hoc simulation. The source has $\pi_s=0.001$ and $Z\mid Y=0\sim N(0,1)$, $Z\mid Y=1\sim N(2.5,1)$. The unperturbed scores are exact source posteriors. Target-prevalence factors were $0.25,0.5,1,2,4$. We crossed negative-class means $-0.30,-0.15,0,0.15,0.30$ with positive-class means $2,2.5,3$ and applied multipliers $0.8,1,1.25$ to the full source logit. We drew two million observations in each of 225 scenarios from one random-number generator (seed 20261103). The floor is half the source prevalence. No observed target label enters the decision.

In the five exact-label-shift scenarios, no case collapsed and the largest optimum error was 9.45 percent. Under violations, the test tracks only the likelihood. Shifting only negative-class scores upward leaves all ten optima above the floor and 221 to 10,837 percent too high. A logit multiplier of 0.8 alone inflates all five optima at least 33-fold. Across the grid, 137 optima exceed the floor, 106 of them more than 100 percent too high. Of the 88 remaining cases, 44 have target prevalence at least twice the floor. These frequencies describe one realization per setting. The mathematical supplement gives the population means for the distorted scores.

\subsection{Delayed-label anchoring}
Two post hoc comparators use the same 30 columns and development evaluation rows as frozen LightGBM, the per-record model (Table~\ref{tab:comparators}). Logistic regression fits the original window with all positives and 2 percent of negatives. Refreshed LightGBM refits through $m-1$ with fixed hyperparameters. Both use the original shape-map and anchor recipe. Logistic regression reaches 4.0 times the aggregate's PR-AUC (frozen 3.9, refreshed 4.9) and beats it on PR-AUC and log loss in all six months, but both cluster intervals for each difference from frozen LightGBM include zero (empirical supplement). Refitting improves pooled PR-AUC and log loss under both partitions but not mean monthly level error. Training and calibration windows overlap. The aggregate gives zero hazard to cells without an event in the carried month. At the log-loss clipping floor of $10^{-12}$, the 165 positives in those cells account for 31 percent of its loss gap to frozen LightGBM. Excluding them preserves the direction. The one such holdout positive accounts for 3 percent of the holdout log-loss gap.

\begin{table}[htbp]
\caption{Per-record models and the aggregate on six registry development months (18,174,176 rows). Level is mean absolute monthly rate bias (\%). The three per-record models use Venn--Abers calibration. All four predictors use the anchor. Table~\ref{tab:ranking-and-calibration} instead scores raw predictions.}
\label{tab:comparators}
\footnotesize
\setlength{\tabcolsep}{4pt}
\begin{tabular}{lrrrr}
\toprule
Model & PR-AUC & Log loss & Brier & Level \\
\midrule
Frozen LightGBM & 0.007270 & 0.008227 & 0.0011703 & 24.2 \\
Refreshed LightGBM & 0.009173 & 0.008115 & 0.0011594 & 28.7 \\
Logistic regression & 0.007402 & 0.008337 & 0.0011416 & 31.2 \\
Aggregate & 0.001870 & 0.009028 & 0.0011419 & 24.7 \\
\bottomrule
\end{tabular}
\end{table}

Compared with the original EM level layer, the anchor roughly halves pooled log loss for frozen LightGBM and the aggregate (empirical supplement). Omitting the level layer narrowly wins pooled PR-AUC for both and pooled log loss for the aggregate, but it does not control monthly workload. After February, unanchored model bias ranges from plus 25 to plus 81 percent. With the anchor, every month except March stays within about 25 percent (mean absolute bias 24.2 against 37.3 percent, Table~\ref{tab:anchored}). The comparison holds after isotonic recalibration of the aggregate and, in July, with a shape map fitted outside the ranker's fitting window (empirical supplement).

\begin{table}[htbp]
\caption{Monthly registry rates and prediction bias before and after adjustment. Raw denotes the ranker, +VA its Venn--Abers calibration, and Final the anchored model. The aggregate uses the same anchor.}
\label{tab:anchored}
\footnotesize
\setlength{\tabcolsep}{4pt}
\begin{tabular}{lrrrrrr}
\toprule
 & \multicolumn{2}{c}{Rate (\%)} & \multicolumn{4}{c}{Level bias (\%)} \\
\cmidrule(lr){2-3}\cmidrule(lr){4-7}
Month & Observed & Anchor & Raw & +VA & Final & Aggregate \\
\midrule
2026-02 & 0.1479 & 0.1704 & +0.1 & $-$0.4 & +3.1 & +12.5 \\
2026-03 & 0.1035 & 0.1773 & +118.4 & +80.8 & +96.2 & +67.6 \\
2026-04 & 0.1029 & 0.1457 & +40.2 & +38.4 & +25.0 & +38.1 \\
2026-05 & 0.1046 & 0.1181 & +35.6 & +32.5 & $-$2.4 & +10.7 \\
2026-06 & 0.1073 & 0.1037 & +73.0 & +46.3 & $-$3.4 & $-$5.2 \\
2026-07 & 0.1202 & 0.1049 & +58.6 & +25.3 & $-$15.3 & $-$14.1 \\
\bottomrule
\end{tabular}
\end{table}

For frozen LightGBM against the aggregate, paired PR-AUC and log-loss intervals exclude zero favorably under record bootstrap and both cluster partitions (empirical supplement). In March the anchor was 71.4 percent high. The anchored model bias was plus 96.2 percent in March and minus 2.4 percent by May.

\subsection{Lag, dependence, and window length}
Four registry partitions give design effects of 59 to 199, or 115 to 219 by bootstrap. The bootstrap effects imply 10.8--14.9 percent relative sampling standard deviation for the anchor. We compare the anchor with subsequent observed rates to see how closely it tracks rate changes. Its three-month window spans roughly 9,400 to 15,600 events. One-month, three-month, and exponential (two-month half-life) anchors have similar pooled log loss and PR-AUC but mean absolute bias of 19.9, 24.2, and 27.3 percent. Trend and drift corrections cut March's anchor error from 71.4 to 31.0 and 43.5 percent but overshoot once the rate settles and worsen mean anchor error on Form D and Kickstarter (empirical supplement). The holdout retained the three-month window. \space

\section{Single-Use Holdout Evaluation}\label{sec:holdout}
In one registry holdout, both predictors shared the three-month anchor.

\subsection{Evaluation design}
We fixed both predictors and both criteria before scoring. Files from the first four days had been parsed for counts, including event counts, before the design was final. This four-day overlap rules out a fully untouched holdout.

The 15-day window held 3,129,111{} rows and 1,507{} positives against roughly 1,800 projected from July's rate. The gap was seven binomial standard errors, or about one under the smallest development design effect, 59.

\paragraph{Use of AI tools.} Under the authors' direction, AI coding assistants helped write and run the study's code and check the reported numbers.{}

\subsection{Primary criteria}
The model met both prespecified criteria (Table~\ref{tab:holdout-results}). A constant score gives average precision equal to the observed rate, 0.000482{}. The model reached 8.03 times this value and the aggregate 1.63 times. Post hoc cluster intervals were 3.2 to 8.5 times as wide as the record-level intervals. With predictors fixed, both partitions favored the model on both metrics in every draw.

All analyses in this paragraph are post hoc. A rebuild with an August 5 file cutoff reproduced all 52 earlier monthly panels exactly. Only August labels changed. Excluding the 380 entities with August 1--4 events, the four early days, left 3,128,731 at risk and 1,127 events over 11 days. On this risk set, model PR-AUC was 0.003277 against the aggregate's 0.000647. Rescaling probabilities to 11 days gave log loss 0.002979 against 0.003188. With an isotonic map fitted through July, the aggregate's full-window PR-AUC was 0.000713 and log loss 0.004143, against the model's 0.003865 and 0.003835. Both metrics favored the model in the 11-day window and in every paired cluster draw for both windows. Both also favored the model on August 7--15, the days after the design was final (empirical supplement).

\begin{table}[htbp]
\caption{Registry holdout results. Differences are per-record minus aggregate. The prespecified criteria use record-level intervals.}
\label{tab:holdout-results}
\footnotesize
\setlength{\tabcolsep}{4pt}
\begin{tabular}{lrr}
\toprule
 & Ranking: PR-AUC & Score: log loss \\
\midrule
Model & 0.0038649& 0.0038352\\
Aggregate & 0.0007846& 0.0041680\\
Difference & +0.0030803& $-$0.0003329\\
Record 95\% CI & [+0.0024664, +0.0039137]& [$-$0.0003734, $-$0.0002942]\\
Criterion & Met& Met\\
\addlinespace
Overall & \multicolumn{2}{c}{Both criteria met} \\
\addlinespace
\multicolumn{3}{l}{Cluster 95\% intervals (post hoc sensitivity analysis)} \\
Jurisdiction & [+0.001669, +0.013902] & [$-$0.000532, $-$0.000190] \\
Managing LOU & [+0.001646, +0.010084] & [$-$0.000474, $-$0.000219] \\
\bottomrule
\end{tabular}
\end{table}

\subsection{Secondary diagnostics}\label{sec:secondary-diagnostics}
Table~\ref{tab:holdout-calibration} reports prespecified CORP~\cite{dimitriadis2021stable} and logistic calibration diagnostics. The model's record-level Wald intercept interval ($-$0.7658{}, [$-$1.0141, $-$0.5175]{}) excludes zero. Its slope interval (0.8852{}, [0.8504, 0.9200]{}) excludes one. Under either post hoc cluster partition, the intercept interval includes zero but the slope interval excludes one. The aggregate departs further from calibration (intercept $-$3.9930{}, slope 0.4631{}).

\begin{table}[htbp]
\caption{Calibration, discrimination, and Brier score on the registry holdout. The jointly fitted intercept and slope have separate record-level intervals.}
\label{tab:holdout-calibration}
\footnotesize
\setlength{\tabcolsep}{4pt}
\begin{tabular}{lrr}
\toprule
Quantity & Model & Aggregate \\
\midrule
Log loss: miscalibration & 0.0000281& 0.0001061\\
\quad discrimination & 0.0003531& 0.0000983\\
\quad uncertainty (shared) & \multicolumn{2}{c}{0.0041602} \\
Brier score & 0.0004822& 0.0004815\\
\quad miscalibration & 0.0000018& 0.0000002\\
\quad discrimination & 0.0000010& 0.0000001\\
\quad uncertainty (shared) & \multicolumn{2}{c}{0.0004814} \\
Calibration intercept & $-$0.7658& $-$3.9930\\
\quad 95\% Wald CI & [$-$1.0141, $-$0.5175]& [$-$4.3565, $-$3.6294]\\
Calibration slope & 0.8852& 0.4631\\
\quad 95\% Wald CI & [0.8504, 0.9200]& [0.4156, 0.5107]\\
ECE, 20 bins, diagnostic & 0.0000872& 0.0002749\\
\bottomrule
\end{tabular}
\end{table}

The model's log loss was 0.0003250 below the constant-score benchmark of 0.0041602{}. The aggregate's was slightly above that benchmark. Discrimination accounts for about three quarters of the model-minus-aggregate difference ($-$0.0003329{}). The full-window model Brier score exceeded the original aggregate's by $7.21\times10^{-7}$. All eight cluster intervals for the Brier difference include zero across the two windows, two aggregate variants, and two partitions. Post hoc, a constant forecast at the exposure-adjusted anchor gave Brier 0.00048138 (below both predictors) and log loss 0.0041630 (between them).

The observed rate was 0.000482{}, the model mean was 0.000471{} (minus 2.3 percent{}), and the aggregate mean was 0.000520{} (plus 8.0 percent{}). The $15/31$ transform takes the monthly anchor from 0.001107{} to 0.000536, 11 percent above the observed rate. It assumes a homogeneous within-month hazard, yet days 1--15 held 51--60 percent of each development month's positives against 48--54 percent of its calendar days. Post hoc, the largest file accounted for 10.6 percent of positives and the largest day 14.7 percent. Omitting any one day's positives (negatives kept) left the PR-AUC ratio between 3.8 and 5.4 and both point differences favorable. We did not recompute intervals. These checks cannot isolate operations spanning files or days.

With the fixed source priors, a post hoc check gives $\mu=0.889$ for the model and 0.972 for the aggregate. All four cluster intervals contain one. The fitting-window prior gives 1.010 and 1.295, respectively. With the original EM update, per-record level error would be minus 99.99 percent and log loss 0.00797 against the aggregate's 0.00754, failing the log-loss criterion. The gated rule would have kept the anchor, matching always anchoring in this window.

\subsection{What the holdout establishes}
The per-record model had higher PR-AUC and lower mean log loss than the aggregate. Both paired intervals excluded zero.{} Anchoring preserves ranks. The PR-AUC gap in this one 15-day window was already present before anchoring. Logistic regression showed a similar development gain (Table~\ref{tab:comparators}). This comparison cannot separate covariate, capacity, and refitting effects. The aggregate's Brier point estimate was below the per-record model's in August and every development month. \space

\section{Discussion}\label{sec:conclusion}
Choose a workload floor before inspecting scores. One pass decides whether a nonflat binary likelihood peaks above the floor. A nonpositive $C(t)$ rules that out. With $C(t)>0$, check the estimate against independent information. One supported estimate was 688 percent too high. For the per-record model in all three streams, anchoring and keeping calibrated scores both gave lower mean level error than the gated rule. At the fixed source priors, all registry cluster intervals for $\mu$ included one.

\paragraph{Limitations.} We claim no novel concavity theorem. We evaluated the gated rule post hoc and declared no operational floor prospectively. The one 15-day registry holdout had a single aggregate comparator. Files from its first four days had been parsed for counts, including event counts, before the design was final. A single window cannot show how often the advantage recurs. The aggregate had the lower Brier point estimate, and neither predictor was exactly calibrated (Section~\ref{sec:secondary-diagnostics}). Our resampling holds the shape map and source prior fixed. All three streams use binary outcomes dated from source records. An alternative to EM needs a justified source of prevalence information and a check on later event counts. \space

\begin{acks}
This research was conducted as part of the ``LLM Cognition Toolbox for SMEs'' project, funded by the Luxembourg Ministry of the Economy, in partnership with the crowdfunding analytics platform Wixdom (\url{https://www.wixdom.io/}).
The experiments presented in this paper were carried out using the HPC facilities of the University of Luxembourg~\cite{varrette2022ulhpc} (see \url{https://hpc.uni.lu}).
\end{acks}
\newpage

\pdfbookmark[1]{Ethical Considerations}{ethics}
\section*{Ethical Considerations}
GLEIF publishes the LEI registry under CC0, and SEC Form D is United States government open data. Both concern legal entities, although LEIs can also identify individuals acting in a business capacity, and the registry panel does not exclude them. The Form~D panel uses only the submission, issuer, and offering tables and excludes the related-person names listed elsewhere in the filings. The Kickstarter panel removes creator identities and personal fields. Its raw snapshot is not redistributed because campaign records may remain linkable to public pages and redistribution rights were not established. No entity-level panel, score, or fitted registry scorer is redistributed, so no risk score for an identifiable firm is published. We downloaded the GLEIF files sequentially, with delays between requests and exponential backoff for retries.

The registry outcome combines administrative transitions, dissolutions, and identifier successions. It is not a measure of creditworthiness. Registry-event forecasts could nevertheless be misused in access or service decisions, especially when a shifted probability level is treated as an individual determination. A deployment should report event-count diagnostics alongside rankings and require human review before any adverse decision about an entity. \space

\end{document}